\documentclass{article}

\usepackage[nonatbib,preprint]{neurips_2024}

\usepackage[utf8]{inputenc} 
\usepackage[T1]{fontenc}    
\usepackage[pagebackref,breaklinks,colorlinks,citecolor=teal]{hyperref}
\usepackage{url}            
\usepackage{booktabs}       
\usepackage{amsfonts}       
\usepackage{nicefrac}       
\usepackage{microtype}      
\usepackage{xcolor}         
\usepackage{multirow}
\usepackage[pdftex]{graphicx}
\usepackage{amsmath}
\usepackage{wrapfig}
\usepackage{float}
\usepackage{mathtools, stmaryrd}
\DeclarePairedDelimiterX{\Iintv}[1]{\llbracket}{\rrbracket}{\iintvargs{#1}}
\NewDocumentCommand{\iintvargs}{>{\SplitArgument{1}{,}}m}
{\iintvargsaux#1} %
\NewDocumentCommand{\iintvargsaux}{mm} {#1\mkern1.5mu..\mkern1.5mu#2}

\newcommand{\C}{\mathbf{C}}
\newcommand{\T}{\mathbf{T}}
\newcommand{\F}{\mathbf{F}}
\newcommand{\W}{\mathbf{W}}

\definecolor{darkyellow}{HTML}{fa921f}
\definecolor{darkgreen}{rgb}{0.0, 0.5, 0.0}

\title{Text2CAD: Generating Sequential CAD Models from Beginner-to-Expert Level Text Prompts}

\newcommand{\eg}{\textit{e.g.,}}
\newcommand{\ie}{\textit{i.e.,}}
\newcommand{\twal}{$\operatorname{Text2CAD} \operatorname{w/o} \operatorname{AL}$ }
\newcommand{\gpt}{$\operatorname{GPT-4V}$ }

\begin{document}
\def\thefootnote{*}\footnotetext{Equal Contributions.}
\def\thefootnote{$\dagger$}\footnotetext{Corresponding author (\text{mohammad.khan@dfki.de)}}
\author{
  \begin{tabular}[t]{c}
    Mohammad Sadil Khan$^{*\dagger 1,2,3}$ \\
  \end{tabular} \hspace{1em}
  \begin{tabular}[t]{c}
    Sankalp Sinha$^{*1,2,3}$ \\
  \end{tabular} \hspace{1em}
  \begin{tabular}[t]{c}
    Talha Uddin Sheikh$^{1,2,3}$ \\
  \end{tabular} \vspace{1em} \\
  \begin{tabular}[t]{c}
    \textbf{Didier Stricker}$^{1,2,3}$ \\
  \end{tabular} \hspace{1em}
  \begin{tabular}[t]{c}
    \textbf{Sk Aziz Ali}$^{1,4}$ \\
  \end{tabular} \hspace{1em}
  \begin{tabular}[t]{c}
    \textbf{Muhammad Zeshan Afzal}$^{1,2,3}$ \\
  \end{tabular} \\
  \\
 \begin{tabular}[t]{c}
    $^1$\text{DFKI} \\
  \end{tabular} \hspace{0.2em}
  \begin{tabular}[t]{c}
    $^2$\text{RPTU Kaiserslautern-Landau} \\
  \end{tabular} \hspace{0.2em}
  \begin{tabular}[t]{c}
    $^3$\text{MindGarage} \\
  \end{tabular} \hspace{0.2em}
  \begin{tabular}[t]{c}
    $^4$\text{BITS Pilani, Hyderabad} \\
  \end{tabular}
}


\maketitle
\vspace*{-.7\baselineskip}
\vspace*{-.7\baselineskip}
\begin{abstract}
\vspace*{-.7\baselineskip}
Prototyping complex computer-aided design (CAD) models in modern softwares can be very time-consuming. This is due to the lack of intelligent systems that can quickly generate simpler intermediate parts. We propose Text2CAD, the first AI framework for generating text-to-parametric CAD models using designer-friendly instructions for all skill levels. Furthermore, we introduce a data annotation pipeline for generating text prompts based on natural language instructions for the DeepCAD dataset using Mistral and LLaVA-NeXT. The dataset contains $\sim170$K models and $\sim660$K text annotations, from abstract CAD descriptions (e.g., \textit{generate two concentric cylinders}) to detailed specifications (e.g., \textit{draw two circles with center} $(x,y)$ \textit{and radius} $r_{1}$, $r_{2}$, \textit{and extrude along the normal by} $d$...). Within the Text2CAD framework, we propose an end-to-end transformer-based auto-regressive network to generate parametric CAD models from input texts. We evaluate the performance of our model through a mixture of metrics, including visual quality, parametric precision, and geometrical accuracy. Our proposed framework shows great potential in AI-aided design applications. Project page is available at~\href{https://sadilkhan.github.io/text2cad-project/}{https://sadilkhan.github.io/text2cad-project/}.


\end{abstract}
   
\vspace*{-.8\baselineskip}
\section{Introduction}\label{sec:introduction}
%
\vspace{-0.3cm}
\setlength{\columnsep}{10pt}%
\begin{wrapfigure}{r}{0.39\textwidth} 
    \includegraphics[width=0.37\textwidth, viewport=0 0 430 420]{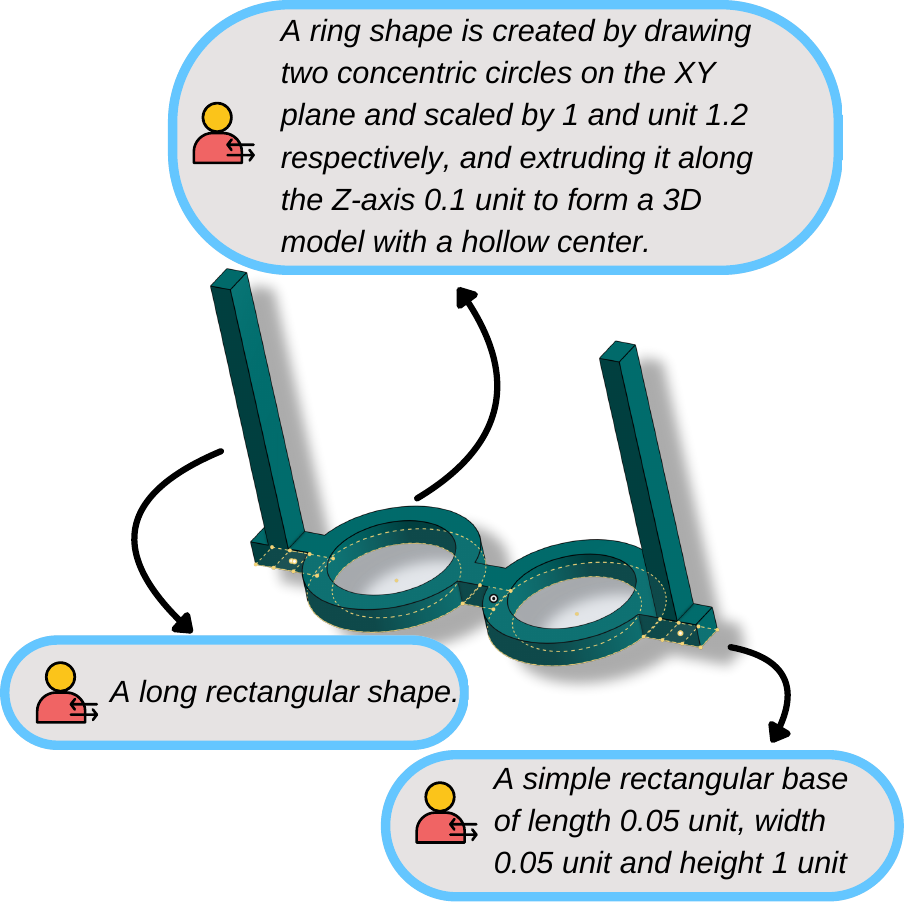}
    \caption{\textit{Designers can efficiently generate parametric CAD models from text prompts. The prompts can vary from abstract shape descriptions to detailed parametric instructions}.}
    \label{fig:teaser}
\end{wrapfigure}
%
%
Computer-Aided Design (CAD) plays a crucial role in industrial design and additive manufacturing (AM), revolutionizing the way products are prototyped~\cite{cherng1998feature}. This type of prototyping requires feature-based part modeling~\cite{cherng1998feature}, precision measurements~\cite{robertson1993cad}, and creative part editing~\cite{YAMAMOTO2005513} at different design stages \cite{robertson1993cad,YAMAMOTO2005513}. While CAD software saves the final model as a boundary representation (B-Rep) \cite{lambourne2021brepnet}, the design process often involves a chain of 2D sketches (e.g., circles, lines, splines) and 3D operations (e.g., extrusion, loft, fillet) \cite{Wu_2021_ICCV,HNCxu2023,xu2022skexgen,mkhan2024cadsignet}. This representation allows the designers to control the design history and iteratively refine the final models. 

Despite their capabilities, modern CAD tools lack the AI-assisted design integration~\cite{REGASSAHUNDE2022100478}. In Figure~\ref{fig:teaser}, we illustrate how an intelligent system capable of generating parametric CAD models from textual descriptions can be utilized to assemble a complex 3D model. Although tools like FreeCAD \cite{freecad}, SolidWorks \cite{solidworks}, and Para-Solid \cite{parasolid} offer 3D CAD models from catalogs like McMaster-Carr~\cite{mcmaster} for the reuse of existing CAD models, no such system currently exists that can generate parametric CAD models from textual design descriptions. One primary challenge for developing such a system is defining suitable textual descriptions for parametric CAD generation, making it difficult to create deep learning methods that accurately convert these descriptions into precise CAD models.

To address this gap, in this paper we propose Text2CAD as the first AI framework for generating parametric CAD models represented by construction sequences (\ie~parameters for 2D sketches and extrusions) from design-related text prompts. We faced two primary challenges in fulfilling this goal: (1) the unavailability of the dataset and (2) a network to map the texts into CAD construction sequences. Towards this end, we introduce a data annotation pipeline to generate a dataset containing textual descriptions of the CAD models in DeepCAD~\cite{Wu_2021_ICCV} dataset. We leverage the open-source Large Language Models (LLMs)~\cite{jiang2023mistral} and Vision Language Models~\cite{liu2023llava,liu2024llava} for this task. Our annotated text prompts are multi-level in nature ranging from highly abstract (\eg~\textit{a long rectangular shape}, \textit{a thin S-shaped object}) to more specific with detailed parametric descriptions (\eg~\textit{first draw a rectangle from ($x_1,y_1$) and ($x_2,y_2$).... then extrude the sketch along z-axis..}). These prompts are designed for users of all skill levels and can contain arithmetic logic and numerical expressions as part of the design details. Within this framework, we introduce Text2CAD Transformer~\cite{transformer}, a conditional deep-generative network for generating CAD construction language~\footnote{In this paper, the phrases \lq CAD construction language \rq, \lq CAD design history \rq and \lq CAD construction sequence \rq are used interchangeably.} from text prompt in an auto-regressive fashion.

Currently, there are works on text-to-3D generation~\cite{lin2023magic3d,nichol2022pointe,wang2023prolificdreamer, mildenhall2021nerf, ho2020denoising} that have shown significant advancements in creating 3D scenes and shapes from textual descriptions. But existing text-to-3D methods are not applicable for generating CAD models from text descriptions as the final output of these models is neither parametric nor human-editable in nature. Very recently web API from \textit{zoo developers}~\cite{zoo} has introduced CAD generation app using text prompt from users and programmable scripting language (as KittiCADLanguage\footnote{\url{https://github.com/KittyCAD/modeling-app/tree/main?tab=readme-ov-file}}) for designers to edit and modify. However, the generated CAD models are obtained in the form of solid-body, and not decomposed to its intermediate \textit{sketch-and-extrusion} steps as proposed in our Text2CAD. On the other hand, Given raw numerical data of any parametric CAD model, current state-of-the-art large language models (LLMs), such as pre-trained Mistral-50b \cite{jiang2023mistral} or GPT-4 \cite{achiam2023gpt} and open source Llama \cite{touvron2023llama} may only derive procedural scripting codes for other APIs, such as FreeCAD \cite{riegel2016freecad} or OpenSCAD \cite{machado2019FreeCAD}, to generate a model. However, in contrast to our Text2CAD, such LLM augmented CAD generation approach will not be designer-friendly, not suitable for beginner-level designers, will not automate the development process in easy ways, and will restrict the re-usability of the scripts in case of complex shapes. Alternatively, Using state-of-the-art vision language models, such as LLaVa \cite{liu2023llava,liu2024llava}, GPT-4V \cite{Zhang2023GPT4V}, as an alternative for deducing CAD construction sequences performs poorly because of two main reasons -- (1) no training datasets are available that provide natural language-based design instructions as annotations for raw CAD construction sequences and (2) most VLMs are trained on categorical description/caption datasets of 3D objects (\eg~LLaVA-NeXT \cite{liu2024llava} predicts \lq two concentric hollow cylinders \rq as \textit{toilet paper}). We remove the above limitations in our Text2CAD by creating new large-scale annotations for DeepCAD \cite{Wu_2021_ICCV} dataset using responses from LLMs and VLMs to train our multi-modal model. Our contributions can be summarized as follows:
\begin{itemize}
    \item We propose Text2CAD as the first AI framework for generating parametric 3D CAD models using textual descriptions.
    \item We introduce a data annotation pipeline that leverages both LLMs and VLMs to generate a dataset that contains text prompts with varying level of complexities and parametric details.
    \item We propose an end-to-end transformer-based autoregressive architecture for generating CAD design history from input text prompts.
    \item Our experimental analysis demonstrates superior performance over the two-stage baseline method adapted for the task at hand.
\end{itemize}

The rest of the sections are organized as follows: Section~\ref{sec:related work} reviews the related work in CAD domains. Section~\ref{sec:text2cad_data_annot} outlines our data annotation pipeline. Section~\ref{sec:text2cad_architecture} details our proposed Text2CAD transformer architecture. Section~\ref{sec:experiment} presents our experimental results. Section~\ref{sec:limitation} discusses the limitations of our current framework, and Section~\ref{sec:conclusion} concludes the paper.

\section{Related Work}\label{sec:related work}
\noindent\textbf{Datasets and Generative models for CAD:} Current datasets and generative models for CAD are limited and often not suited for developing knowledge-based CAD applications. Some datasets focus solely on 2D sketch design \cite{seff2020sketchgraphs,CadAsLangNIPS21,seff2021vitruvion}, and other popular datasets like ABC \cite{koch2019abc}, Fusion360 Gallery \cite{willis2021fusion}, Thingi10K \cite{Zhou2016Thingi10KAD}, and CC3D \cite{cc3d,CADOPsNetDV22} provide 3D meshes, BRep (boundary representation), and other geometry or topology related annotations that are suitable for 3D modeling. DeepCAD \cite{Wu_2021_ICCV} dataset, a subset of ABC, and Fusion360 \cite{willis2021fusion} provide CAD construction sequences in the form of \textit{ sketch and extrusion} to deduce design history. 
However, CAD models may consist of numerous other types of operations beside \textit{extrusion}, and such construction sequences with other CAD operations are not available in the current datasets. Finally, there is no dataset available that provides textual design descriptions as annotations to create a conversational AI system for CAD modeling.


Current supervised learning methods that fall under \textit{sequence-to-sequence Sketch/CAD language modeling} \cite{Wu_2021_ICCV,HNCxu2023,mkhan2024cadsignet,CadAsLangNIPS21} filters out unnecessary metadata from lengthy raw design files and represent them as desired sequence of input/output tokens. For instance, Ganin et al. \cite{CadAsLangNIPS21} represents design files as messages in Protocol Buffer \cite{varda2008google} format. Hierarchical Neural Coding (HNC) method \cite{HNCxu2023} represents the desired design sequence in tree structure of 2D sketch loops, 2D bounding boxes over all loops as profile, and 3D bounding boxes over all profiles as solid. CAD-SIGNet~\cite{mkhan2024cadsignet} represents CAD construction language as a sequence composed of 2D sketch and extrusion parameters. In Text2CAD method, we map the raw design history obtained from DeepCAD metadata into textual descriptions.      


\noindent\textbf{CAD Construction Language using Transformers:}
Transformer-based \cite{transformer} network architecture is the preferred choice for many deep learning-based applications related to CAD modeling \cite{Wu_2021_ICCV}, 3D scan-to-CAD reverse engineering \cite{mkhan2024cadsignet,CADGenRoundedVox2022}, representation learning \cite{jung2024contrastcad} and others \cite{ritchie2023neurosymbolic}. CAD as a language \cite{CadAsLangNIPS21} describe how 2D sketches can be transformed into \textit{design language} by sequencing tokens of 2D parametric curves as message passing units. Mixture of Transformer \cite{transformer} and Pointer Networks \cite{vinyals2015pointer} decode the sketch parameters in auto-regressive fashion.

Formalizing constrained 2D sketches, \ie~collection of curves (\eg~\textit{line, arc, circle and splines}) with dimensional and geometric constraints (\eg~\textit{co-incidence, perpendicular, co-linearity}), as a language for CAD modeling has been studied over last few years \cite{NEURIPS2021SketchGen,CadAsLangNIPS21,nash2020polygen,willis2021engineering,li2022free2cad}. However, the first proposal of developing a CAD language interface was suggested decades ago in \cite{samad1986natural}. Among the recent works in this direction, SketchGen \cite{NEURIPS2021SketchGen} represents 2D sketches as a sequence of the tokens for curves and constraints. The decoder-only transformer model in \cite{NEURIPS2021SketchGen} predicts optimal sketches through nucleus sampling \cite{Holtzman2020NucleusSampling} of token embedding vectors, focusing on replicating drawing processes of CAD designers. Polygen \cite{nash2020polygen} method also employs Transformer model \cite{transformer} to generate detailed 3D polygonal meshes by learning joint distribution on vertices and faces of a CAD. As an extension of  \cite{nash2020polygen}, TurtleGen \cite{willis2021engineering} also propose decoder-only transformer model to learn joint distribution of vertices and edges together that form sketches and represented as graphs in CAD models. 

3D CAD modeling steps as a language is not directly formulated by any state-of-the-art multi-modal CAD learning methods \cite{Wu_2021_ICCV,multicad,mkhan2024cadsignet,xu2022skexgen,CADOPsNetDV22,mallis2023sharp,HNCxu2023,li2022free2cad}. Khan et al. \cite{mkhan2024cadsignet} propose a novel auto-regressive generation of \textit{sketch-and-extrusion} parameters directly from 3D point clouds as input whereas DeepCAD \cite{Wu_2021_ICCV}, SkexGen \cite{xu2022skexgen}, HNC \cite{HNCxu2023} and MultiCAD \cite{multicad} adopts a two-stage strategy to generate the output. MultiCAD~\cite{multicad} adopt multi-modal contrastive learning to associate geometry features with features of CAD construction sequences whereas CAD-SIGNet \cite{mkhan2024cadsignet} requires an extra step as user feedback to vote for one of the many generated sketches at current step to predict the next one. Unlike previous approaches, our proposed Text2CAD transformer is the first auto-regressive network that generates CAD construction sequences directly from textual descriptions.

\vspace*{-.7\baselineskip}
\section{Text2CAD Data Annotation}\label{sec:text2cad_data_annot}

\begin{figure}
    \centering
    \includegraphics[width=1.0\linewidth]{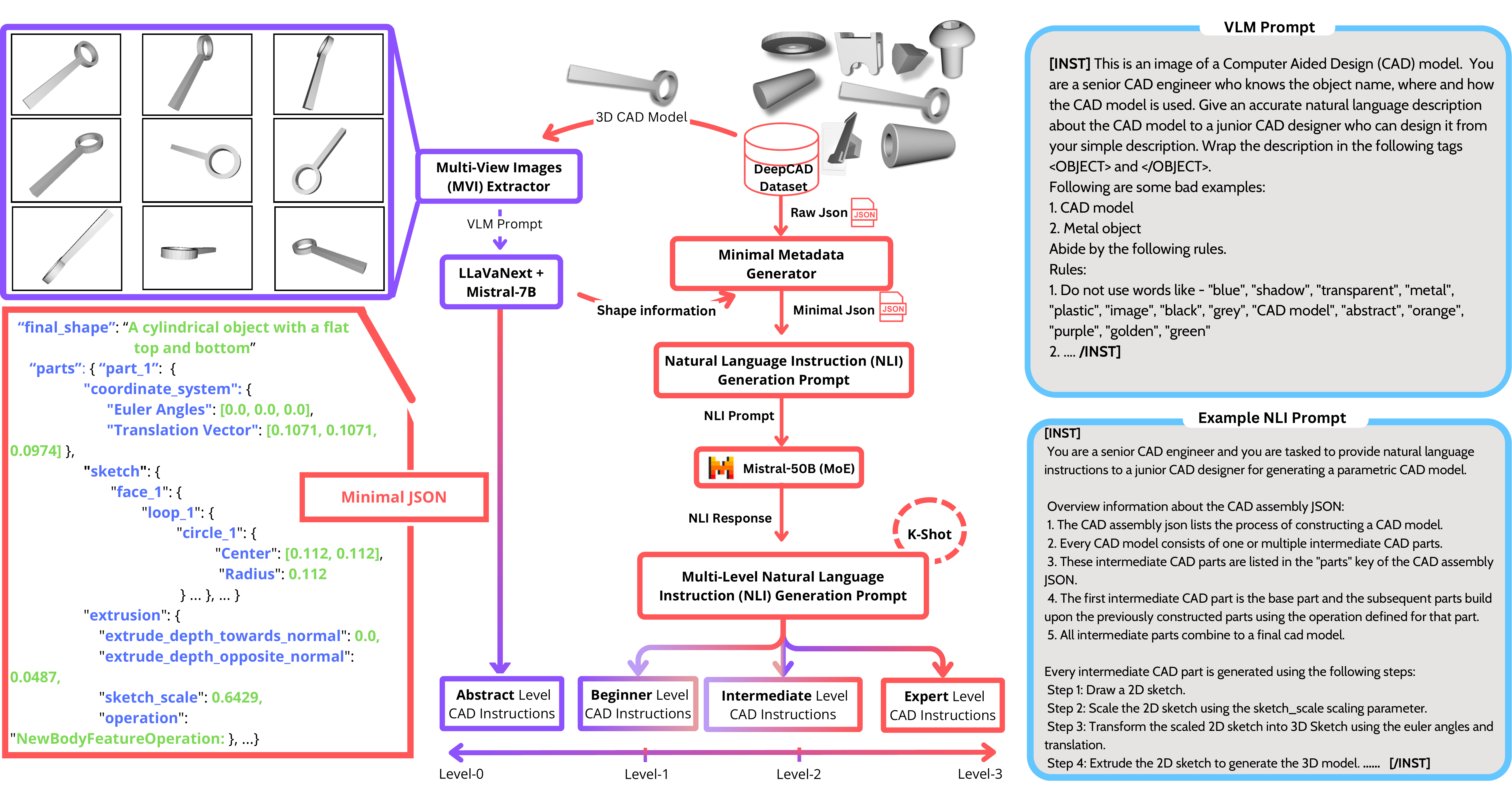}
    \vspace*{-0.5cm}
    \caption{\textbf{Text2CAD Data Annotation Pipeline:} Our data annotation pipeline generates multi-level text prompts describing the construction workflow of a CAD model with varying complexities. We use a two-stage method - (Stage 1) Shape description generation using VLM (Stage 2) Multi-Level textual annotation generation using LLM.}
    \vspace*{-.8\baselineskip}
    \label{fig:LLM-based-DataGen}
        \vspace*{-.8\baselineskip}
\end{figure}
    \vspace*{-.7\baselineskip}
The diagram in Fig.~\ref{fig:LLM-based-DataGen} outlines the process of generating textual annotations for DeepCAD dataset \cite{Wu_2021_ICCV} using Large Language Models (LLMs) \cite{jiang2024mixtral,achiam2023gpt,touvron2023llama} and Vision Language Models (VLMs) \cite{liu2023llava,liu2024llava}. These annotations describe the corresponding CAD construction workflow in human interpretable format.
%
%
To enrich the DeepCAD~\cite{Wu_2021_ICCV} dataset with textual annotations, we implement a two-stage description generation pipeline using the capabilities of both LLMs and VLMs. The two stages are - (1) generating abstract shape descriptions using VLM, and (2) extracting multi-level textual instructions from LLM based on the shape descriptions and design details provided in the dataset. An example text prompt for the CAD model shown in \textit{top-left} of the Figure~\ref{fig:LLM-based-DataGen}: \lq\textit{The CAD model consists of a \textcolor{violet}{cylindrical object with a flat top and bottom} connected by a curved surface and slightly tapered towards the bottom. This object is created by first setting up a coordinate system, then sketching two concentric circles and drawing a closed loop with lines and an arc on a shared plane. The sketch is then extruded along the normal direction to form a solid body. The resulting part has a height of approximately 0.0487 units} \rq. In this example, the phrase in the violet color is generated by a VLM. An LLM uses this description along with the CAD construction information to generate the prompt.

\noindent \textbf{Shape Descriptions using VLM:} The initial step of our annotation generation pipeline involves generating abstract object-level descriptions of the CAD models using LLaVA-NeXT~\cite{liu2024llava} model. The objective in this step is to accurately capture the structural descriptions of the 3D shape, such as \textit{"a ring-like structure"}, \textit{"a cylinder"}, or \textit{"a hexagon with a cylinder on top"}. 
We generate shape descriptions for both the final CAD model and its intermediate parts. We first produce multi-view images from predetermined camera angles for each individual parts and the final CAD model. These images are then utilized in a predefined prompt (refer to the \textit{top-right} of Figure~\ref{fig:LLM-based-DataGen}) for the LLaVA-NeXT~\cite{liu2024llava} model to generate simplified shape descriptions of all individual parts as well as the complete final shape.



\noindent \textbf{Multi-level Design Instructions using LLM:} In this stage, multiple textual annotations corresponding to different design details of a CAD model are generated using Mixtral-50B~\cite{jiang2024mixtral} through a series of steps (refer to the \textit{middle-column} in Figure~\ref{fig:LLM-based-DataGen}). The DeepCAD~\cite{Wu_2021_ICCV} dataset contains CAD construction sequences in JSON format. We first preprocess the raw CAD construction sequences using a \lq Minimal Metadata Generator \rq which replaces random, meaningless keys with more meaningful terms (\eg "part\_1", "loop\_1"). This step aims to reduce the hallucinations~\cite{llm_hallucinations} by Mixtral-50B~\cite{jiang2024mixtral}. The minimal metadata is further augmented with the shape descriptions for each parts and the final model generated by the VLM. The output of this process is a condensed representation of the shapes and their relational attributes within the CAD design (see \textit{bottom-left} in Figure~\ref{fig:LLM-based-DataGen}). With the minimal metadata at hand, we then craft a prompt (refer to the \textit{bottom-right} in Figure~\ref{fig:LLM-based-DataGen}) to generate detailed natural language instructions (NLI) ensuring a minimal loss of information from the minimal metadata. Afterward, the NLI responses are refined by LLM using a $k$-shot~\cite{k_shot_prompting} "Multi-Level Natural Language Instruction Generation Prompt" to generate multi-level instructions of different specificity and details. We categorize these levels as:
\begin{itemize}  
 \vspace*{-.5\baselineskip}
\item\textbf{Abstract level (L0):} Abstract Shape Descriptions of the final CAD model extracted using VLM in the first stage.

\item \textbf{Beginner level (L1):} Simplified Description - Aimed at laypersons or preliminary design stages, this level provides a simplified account of the design steps, eschewing complex measurements and jargon.

\item\textbf{Intermediate level (L2):} Generalized Geometric Description - This level abstracts some of the details, providing a generalized description that balances comprehensibility with technical accuracy.

\item\textbf{Expert level (L3):} Detailed Geometric Description with Relative Values - Here, the instructions include precise geometric descriptions and relative measurements, catering to users who require an in-depth understanding or are performing the CAD modeling task. 
 \vspace*{-.5\baselineskip}
\end{itemize}

Our annotations consist of the generated multi-level instructions at the final stage. We generate these annotations over the course of $10$ days. It's worth noting that one can directly generate the multi-level instructions from the minimal metadata without creating the detailed natural language instructions in the second stage. We observe that this strategy increases the LLM's tendency for hallucinations~\cite{llm_hallucinations} and it generates more inaccurate multi-level instructions. Instead our method follows \textit{chain-of-thought} prompting strategy as outlined in~\cite{wei2022chain} which greatly reduces such bottleneck. More details on our annotation pipeline are provided in Section~\ref{sec:additional_details_data_annot} and~\ref{sec:additional_details_data_annot_2} of the supplementary material.
 




\vspace*{-.7\baselineskip}
\section{Text2CAD Transformer} \label{sec:text2cad_architecture}
\begin{figure}[t]
    \vspace*{-.6\baselineskip}
    \centering
     \includegraphics[width=1.0\linewidth]{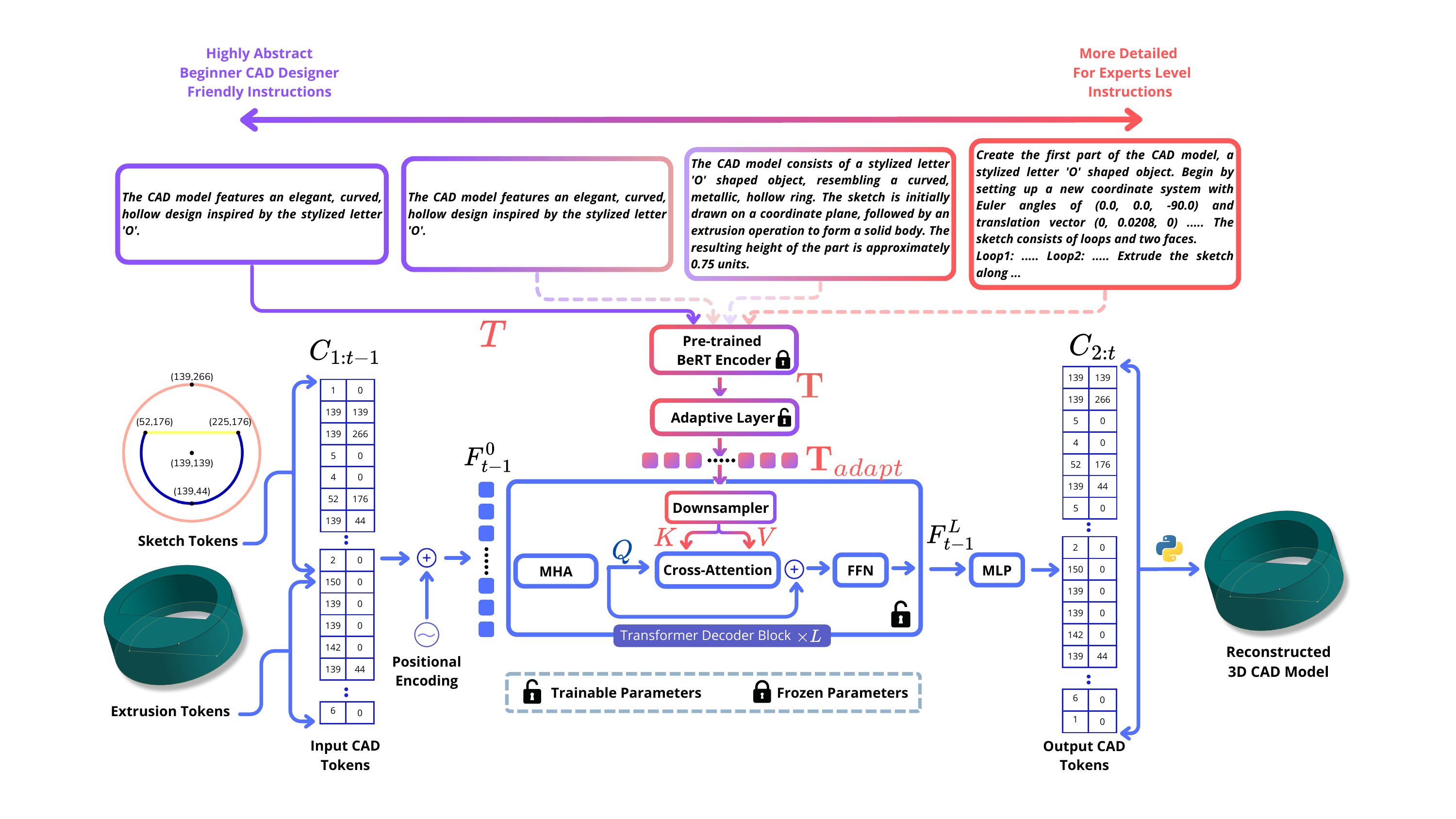}
      \vspace*{-0.7cm}
    \caption{\textbf{Network architecture}: Text2CAD Transformer takes as input a text prompt $T$ and a CAD subsequence $\mathbf{C}_{1:t-1}$ of length ${t-1}$. The text embedding $\T_{adapt}$ is extracted from $T$ using a pretrained BeRT Encoder (~\cite{Bert}) followed by a trainable Adaptive layer. The resulting embedding $\T_{adapt}$ and the CAD sequence embedding $\F^0_{t-1}$ is passed through $\mathbf{L}$ decoder blocks to generate the full CAD sequence in auto-regressive way.}
    \label{fig:architecture}
    \vspace*{-.8\baselineskip}
\end{figure}

\vspace*{-.4\baselineskip}
The Text2CAD transformer architecture, as shown in Figure \ref{fig:architecture}, is designed to transform natural language descriptions into 3D CAD models by deducing all its intermediate design steps autoregressively.
Given an input text prompt $T \in \mathbb{R}^{N_p}$, where $N_p$ is the number of words in the text, our model learns the probability distribution, $P(\C|T)$ defined as
\vspace*{-.5\baselineskip}
\begin{equation}
\vspace*{-.3\baselineskip}
    P(\mathbf{C}|T)=\prod_{t=1}^{N_c} P(c_t|c_{1:t-1}, T;\theta)
\end{equation}
where $\C$ is the output CAD sequence, $N_c$ is the number of tokens in $\C$ and $\theta$ is the learnable model parameter. We represent $\C$ as a sequence of sketch and extrusion tokens as proposed in~\cite{mkhan2024cadsignet}. Each token $c_t \in \C$ is a 2D token that either denotes a (1) 2D-coordinate of the primitives in sketch, (2) one of the extrusion parameters (euler angles/translation vector/extrusion distances/boolean operation/sketch scale) or (3) one of the end tokens (curve/loop/face/sketch/extrusion/start sequence/end sequence). Following~\cite{Wu_2021_ICCV, mkhan2024cadsignet}. We quantize the 2D coordinates as well as the continuous extrusion parameters in $8$ bits resulting in $256$ class labels for each token. An example CAD sequence representation is provided in Figure~\ref{fig:architecture} (in blue table). For more details, please refer to the supplementary section~\ref{sec:cad_seq_rep}. 

\noindent Now we elaborate on the various components of the architecture, detailing the processes involved in converting text to CAD representations. Let the input text prompt at timestep $t-1$ be $T \in \mathbb{R}^{N_p}$ and the input CAD subsequence $\C_{1:t-1} \in \mathbb{R}^{N_{t-1} \times 2}$.

\noindent\textbf{Pretrained Bert Encoder:} 
The initial step in the Text2CAD network involves encoding the textual description provided by the user. This description can range from highly abstract, beginner-friendly instructions to detailed, expert-level commands. To handle this diversity, we used a pre-trained BERT (Bidirectional Encoder Representations from Transformers)~\cite{Bert} model, denoted $\text{BERT}_{\text{pre-trained}}$. The input text $T \in \mathbb{R}^{N_t}$ is tokenized and passed through the BERT model to generate contextual embedding:
\vspace*{-.5\baselineskip}
\begin{equation}
\mathbf{T} = \text{BERT}_{\text{pre-trained}}(T)
\end{equation}
Here, $\mathbf{T} \in \mathbb{R}^{N_p \times d_p}$ represents the sequence of token embedding vectors that capture the semantic meaning of the input text, where $N_p$ is the number of tokens and $d_p$ is the dimension of the embedding.

\noindent\textbf{Adaptive Layer. } An adaptive layer consisting of $1$ transformer encoder layer, refines the output $\T$ of the BERT encoder to better suit the CAD domain aligning with the specific vocabulary and structural requirements of CAD instructions. The adaptive layer outputs the embedding $\mathbf{T}_{\text{adapt}} \in \mathbb{R}^{N_p \times d_p}$ using
\vspace*{-.6\baselineskip}
\begin{equation}
\vspace*{-.3\baselineskip}
\mathbf{T}_{\text{adapt}} = \text{AdaptiveLayer}(\mathbf{T})
\end{equation}

%
%
\noindent\textbf{CAD Sequence Embedder:} Each token in the input CAD subsequence $\mathbf{C}_{1:t-1}$ is initially represented as a one-hot vector with a dimension of 256, resulting in a one-hot representation, $\mathbf{C}^o_{1:t-1} \in \mathbb{R}^{N_{t-1} \times 2 \times 256}$. For the sake of simplicity, we represent $\mathbf{C}^o_{1:t-1}=[\C^{ox}_{1:t-1};\C^{oy}_{1:t-1}]$, where $\C^{ox}_{1:t-1},\C^{oy}_{1:t-1} \in \mathbb{R}^{N_{t-1} \times 256}$. The initial CAD sequence embedding $\F^0_{t-1} \in \mathbb{R}^{N_{t-1} \times d}$ is obtained using Eq.~\ref{eq:cad_embedder}
\begin{equation}\label{eq:cad_embedder}
\F^0_{t-1} = \C^{ox}_{1:t-1}\W^x_{t-1} + \C^{oy}_{1:t-1}\W^y_{t-1} + \mathbf{P}
\end{equation},
where $\W^x_{t-1}, \W^y_{t-1} \in \mathbb{R}^{N_{t-1} \times d}$ are learnable weights and $\mathbf{P} \in \mathbb{R}^{N_{t-1} \times d}$ is the positional encoding.


\noindent\textbf{Layer-wise Cross Attention.} We use a standard transformer decoder~\cite{transformer} with layer-wise cross-attention mechanism between the CAD and the text embedding within the decoder blocks. The layerwise cross-attention mechanism facilitates the integration of contextual text features with the CAD embedding, allowing the model to focus on relevant parts of the text during CAD construction. Each decoder block $l$ takes as input CAD embedding $\F^{l-1}_{t-1}$ and text embedding $\T_\text{adapt}$, where $\F^{l-1}_{t-1}$ is the output of the previous decoder block (for the first decoder block, the input CAD embedding is $\F^0_{t-1}$).
At first, the CAD embedding $\F^{l}_{t-1} \in \mathbb{R}^{N_{t-1} \times d}$ is generated from $\F^{l-1}_{t-1}$ using
\vspace*{-.6\baselineskip}
\begin{equation}
\vspace*{-.2\baselineskip}
    \F^{l}_{t-1} = \operatorname{MHA}(\F^{l-1}_{t-1})
\end{equation}
, where $\operatorname{MHA}$ is the multi-head self-attention~\cite{transformer} operation. Afterwards, We downsample $\T_\text{adapt}$ to generate $\T^l_\text{adapt} \in \mathbb{R}^{N_p \times d}$ using
\vspace*{-.3\baselineskip}
\begin{equation}
\vspace*{-.2\baselineskip}
    \T^l_\text{adapt}=\T_\text{adapt}\W^l_\text{adapt}
\end{equation},
where $\W^l_\text{adapt} \in \mathbb{R}^{d_p \times d}$ is the learnable projection matrix. The cross-attention mechanism involves query ($\mathbf{Q}$), key ($\mathbf{K}$), and value ($\mathbf{V}$) generation using
\vspace*{-.3\baselineskip}
\begin{equation}
\vspace*{-.3\baselineskip}
\mathbf{Q} = \F^{l}_{t-1} \mathbf{W}_Q, \quad \mathbf{K} =\mathbf{T}^l_{\text{adapt}} \mathbf{W}_K, \quad \mathbf{V} = \mathbf{T}^l_{\text{adapt}}\mathbf{W}_V
\end{equation}
Here, $\mathbf{W}_Q \in \mathbb{R}^{d \times d_q}$, $\mathbf{W}_K \in \mathbb{R}^{d \times d_k}$, and $\mathbf{W}_V \in \mathbb{R}^{d \times d_v}$ are learned projection matrices. The cross-attention output $\mathbf{A} \in \mathbb{R}^{N_{t-1} \times d_v}$ is computed as:
\begin{equation}
\mathbf{A} = \operatorname{Softmax}\left(\frac{\mathbf{Q} \mathbf{K}^\top}{\sqrt{d_k}}\right) \mathbf{V}
\end{equation}
%
where $d_k$ is the dimensionality of the key vectors. The cross attention mechanism enables the model to dynamically adjust the importance of different parts of the text relative to the CAD sequence. Afterwards, the output embedding of the decoder block $l$ is generated using
\vspace*{-.4\baselineskip}
\begin{equation}
     \F^{l}_{t-1} \leftarrow \operatorname{LayerNorm}(\operatorname{FFN}(\operatorname{LayerNorm}(\F^{l}_{t-1} + \operatorname{Dropout(A)})))
\end{equation}
,where $\operatorname{FFN}$ is the feed foward network~\cite{transformer} and $\operatorname{LayerNorm}$ is the Layer Normalization~\cite{transformer}.
%
%
The complete Transformer decoder block is repeated $L$ times, allowing for deeper integration and refinement of the text and CAD tokens. The final CAD embedding $F^L_{t-1} \in \mathbb{R}^{N_{t-1} \times d}$ is passed to an $\operatorname{MLP}$ to generate the output CAD sequence. We use Cross-Entropy loss during training. 
\vspace*{-.8\baselineskip}
\section{Experiment}\label{sec:experiment}
\vspace*{-.8\baselineskip}



\begin{figure}
    \centering
    \includegraphics[width=1.0\linewidth]{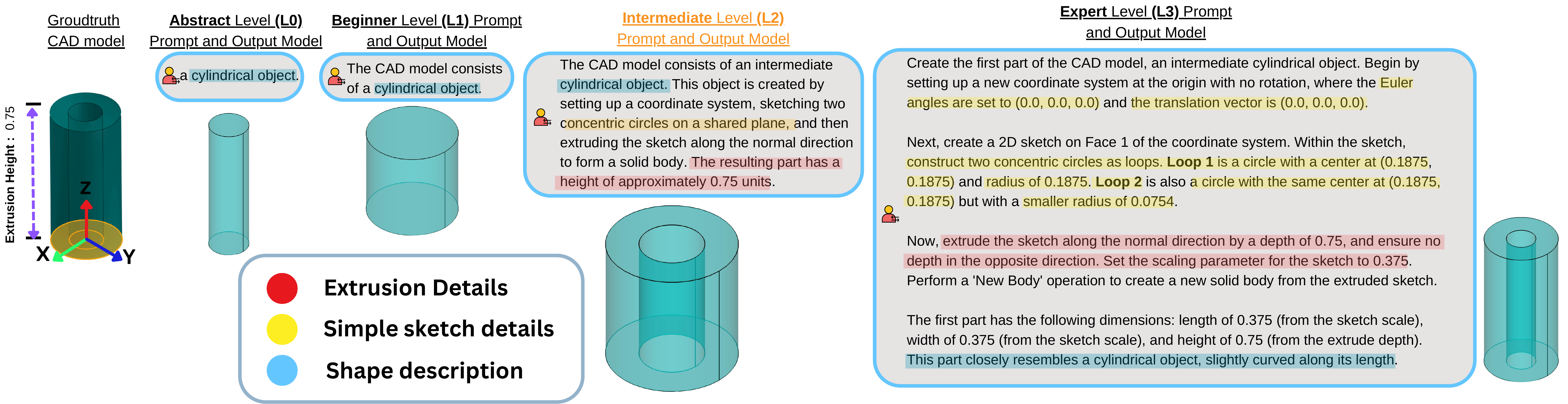}
    \vspace*{-.9\baselineskip}
    \caption{Parametric CAD model generation by Text2CAD transformer using different text prompts. Our text prompts follow a certain structure highlighting the different design aspects of CAD construction workflow (shown in different colors).}
    \vspace*{-.8\baselineskip}
    \label{fig:ModelPredictions_textSelection}
    \vspace*{-.5\baselineskip}
\end{figure}
\noindent \textbf{Dataset. } We use the DeepCAD~\cite{Wu_2021_ICCV} dataset which contains approximately~$\sim150$k training CAD sequences and $\sim8$k test and validation sequences in sketch-and-extrude format. Following, ~\cite{Wu_2021_ICCV, mkhan2024cadsignet}, the sketches and the final CAD models are normalized within a unit bounding box. For each sample in the dataset, four design prompts ranging from abstract to expert levels (\textbf{L0, L1, L2, L3}) are generated using our data annotation pipeline resulting in $\sim600$k training samples, and $\sim32$k test and validation samples.  

\noindent \textbf{Implementation Details. } Text2CAD transformer consists of $L=8$ decoder blocks with $8$ self-attention heads. The learning rate is $0.001$ with AdamW~\cite{adamw} optimizer. Dropout is $0.1$. Maximum number of word tokens, $N_p$ is fixed as $512$ and CAD tokens $N_c$ as $272$. The dimension $d_p$ for the pre-trained Bert encoder~\cite{Bert} embedding $\T$ as well as $\T_{adapt}$ is $1024$. The CAD sequence embedding $d$ is $256$. Following~\cite{mkhan2024cadsignet}, the first two decoder blocks do not use any cross-attention operation between the text embedding and the CAD sequence embedding. The Text2CAD transformer has been trained with teacher-forcing~\cite{teacher_forcing} strategy for $160$ epochs using $1$ Nvidia A$100$-$80$GB GPU for $2$ days. During inference, top-$1$ sampling has been used to autoregressively generate the CAD sequences from an input text.

\noindent \textbf{Baseline. } Since there are no existing methods for generating parametric CAD sequences from text prompts, we use DeepCAD~\cite{Wu_2021_ICCV} and our $\operatorname{Text2CAD} \operatorname{w/o} \operatorname{AL}$ (\ie~without Adaptive Layer) variant as our baselines. To adjust DeepCAD for performing CAD generation from text inputs, the Adaptive Layer~\cite{Wu_2021_ICCV}  (see Section \ref{sec:text2cad_architecture}) is trained to map the pre-trained Bert embedding $\T$ into the ground truth latent vector $z$. During inference, the predicted latent vector $z$ is then passed to the pre-trained DeepCAD decoder to generate the CAD sequences. For $\operatorname{Text2CAD} \operatorname{w/o} \operatorname{AL}$, the pre-trained Bert embedding $\T$ is directly passed to the transformer decoder.

\noindent\textbf{Experimental Setup. }
For setting up our evaluation protocol, the selection of the input text prompts and desired outcomes of the CAD models are depicted in Figure \ref{fig:ModelPredictions_textSelection}. 
Our textual annotations follow a certain structure. In the abstract (L0) and beginner (L1) level prompts, the shape descriptions are more prevalent (in \textcolor{teal}{teal color} in Figure~\ref{fig:ModelPredictions_textSelection}). The intermediate level prompts augment simple sketch (in \textcolor{darkyellow}{yellow color}) and extrusion (in \textcolor{red}{red color)} details with the shape descriptions. Expert-level prompts include more precise details for each of the design aspects previously highlighted. In all our experiments discussed below, we use these four levels of prompts following the aforementioned formats. However, we have conducted another experiment, where we interpolate between the abstract and expert prompts to generate multiple new prompts and observe the performance of our model on these prompts. Due to the space restriction, we provide the results of this experiment in the supplementary Section~\ref{sec:interpolation}.
\vspace*{-.6\baselineskip}
\subsection{Evaluation}
\vspace*{-.6\baselineskip}
Our goal is to evaluate how well the generated CAD sequences align with their respective input text prompts. We concentrate on two primary aspects: (1) examining the \textit{parametric correspondence} between the predicted CAD sequences and the input text prompts, and (2) conducting \textit{visual inspections} of the reconstructed 3D CAD models. Currently, there is no standardized benchmark for text-to-3D models that can be directly applied to this task. Existing methods for text-to-3D~\cite{wang2023prolificdreamer, Jain2021ZeroShotTO} utilize pre-trained CLIP~\cite{sanghi2023clip} models to measure the alignment between reconstructed 3D models and their text prompts. However, CLIP~\cite{sanghi2023clip} scores may not adequately evaluate the geometric~\cite{wu2023gpteval3d} and the parametric aspects of CAD designs. Therefore, inspired by~\cite{wu2023gpteval3d, he2023t3bench}, we employ three evaluation metrics to thoroughly measure the parametric and geometric alignment between the generated CAD designs and the text prompts.


\begin{figure}
    \centering
    \vspace*{-.8\baselineskip}
    \includegraphics[width=1.0\linewidth]{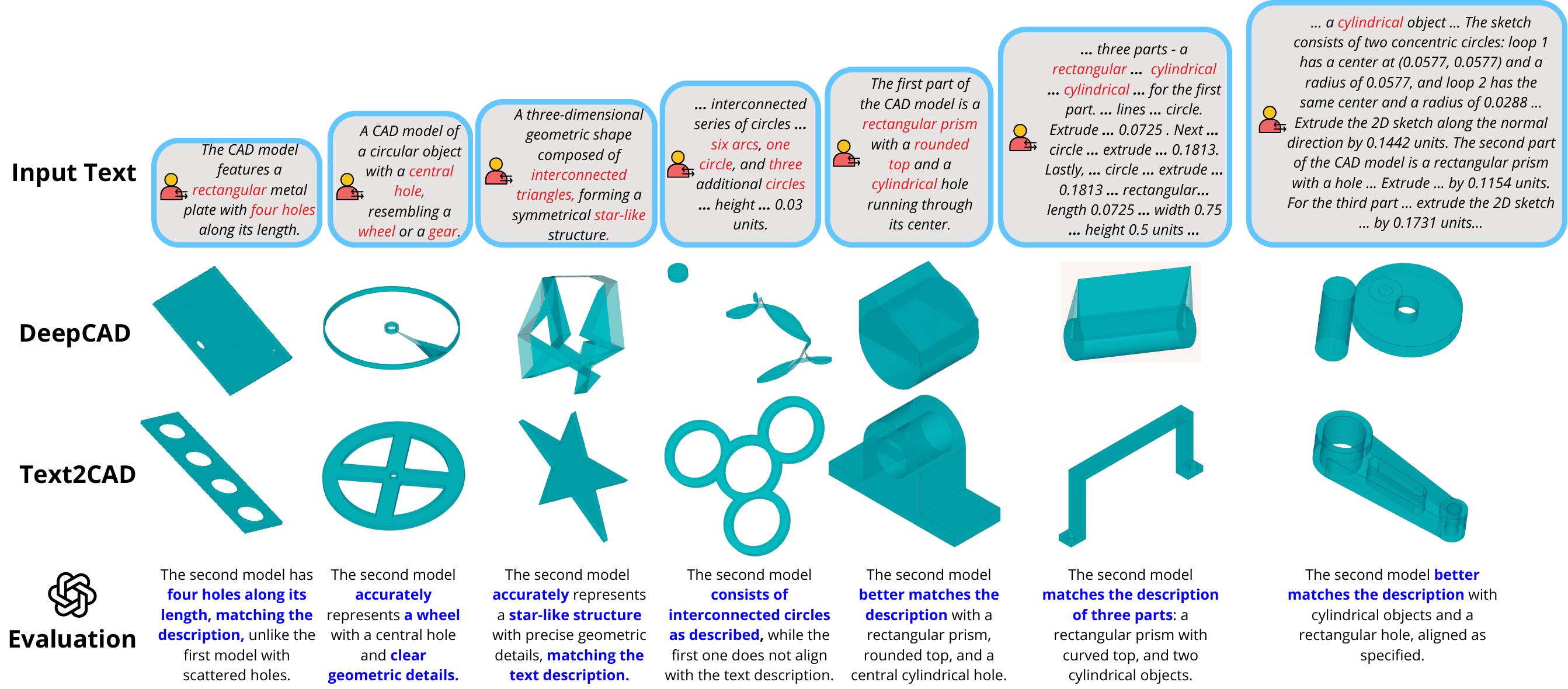}
     \vspace*{-.8\baselineskip}
    \caption{Qualitative results of the reconstructed CAD models of DeepCAD~\cite{Wu_2021_ICCV} and Text2CAD on DeepCAD~\cite{Wu_2021_ICCV} dataset. From top to bottom - Input Texts, Reconstructed CAD models using DeepCAD and Text2CAD respectively and \gpt Evaluation.}
    \label{fig:qual_eval}
\end{figure}
\noindent \textbf{A. CAD Sequence Evaluation:} In this evaluation strategy, we comprehensively assess the \textit{parametric correspondence} between the generated CAD sequences with the input text. We use the groud truth CAD sequence for this purpose. We only use this evaluation strategy for expert-level (L3) prompts. Since, expert-level prompts, being highly detailed, exhibit a higher one-to-one correspondence with the ground truth CAD construction sequences compared to other levels.

\noindent To measure the correspondence between the ground truth and the predicted sequence, the strategy outlined in~\cite{mkhan2024cadsignet} is followed. We evaluate the F1 scores of the primitives and extrusions by aligning the predicted loops with the ground truth loops within the same sketch using the Hungarian matching algorithm~\cite{hungarian-matching} (An example is provided in supplementary Figure~\ref{fig:primitive_loop_matching}). The geometric alignment between the ground truth and reconstructed CAD models is measured using the \textit{chamfer distance} (CD). The \textit{Invalidity Ratio} (IR) is calculated to measure the proportion of invalid CAD sequences.

\begin{table}[ht]
\vspace*{.6\baselineskip}
\caption{Quantitative evaluation between DeepCAD~\cite{Wu_2021_ICCV} and our method (AL is Adaptive Layer). The scores are evaluated only for \textbf{Expert Level} (L3) prompts. The results include F1 scores for primitives and extrusions as well as mean and median CD and IR. CD is multiplied by $10^3$.}
    \label{tab:full_eval}
\centering
\resizebox{\linewidth}{!}{
\begin{tabular}{lcccc|ccc}
\hline
\multicolumn{1}{c}{\multirow{2}{*}{Model}} & \multicolumn{4}{c|}{F1$\uparrow$}                                                                                          & \multirow{2}{*}{\begin{tabular}[c]{@{}c@{}}Median\\ CD$\downarrow$\end{tabular}} & \multirow{2}{*}{\begin{tabular}[c]{@{}c@{}}Mean\\ CD$\downarrow$\end{tabular}} & \multirow{2}{*}{IR$\downarrow$} \\
\multicolumn{1}{c}{}                       & \multicolumn{1}{l}{Line} & \multicolumn{1}{l}{Arc} & \multicolumn{1}{l}{Circle} & \multicolumn{1}{l|}{Extrusion} &                                                                      &                                                                    &                     \\ \hline
DeepCAD                                    & 76.78                    & 20.04                   & 65.14                      & 88.72                          & 32.82                                                                & 97.93                                                              & 10.00               \\
Text2CAD w/o AL                            & 78.88                    & 27.18                   & 71.44                      & 93.28                          & 0.82                                                                 & 35.91                                                              & 2.69                \\
\textbf{Text2CAD}                            & \textbf{81.13}                    & \textbf{36.03 }                  & \textbf{74.25}                      & \textbf{93.31}                          & \textbf{0.37}                                                                 & \textbf{26.41}                                                              & \textbf{0.93}   \\ \hline            
\end{tabular}
}
\label{tab:quan_eval}
\vspace*{.4\baselineskip}
\end{table}

\noindent Table~\ref{tab:quan_eval} summarizes the quantitative results between the baseline methods and our final Text2CAD transformer. Compared to the baselines (rows 1-2), our model (row 3) achieves higher F1 scores for all the primitives and extrusion, with the most notable being an $80\%$ improvement in the F1 score for arcs. The results indicate a better correspondence between the expert-level prompts and the predicted CAD sequences. Notably, our model significantly outperforms the baseline DeepCAD in terms of median CD and invalidity ratio by a factor of $\sim 88.7$ and $\sim10.75$ respectively. The higher CD despite having relatively high F1 scores for DeepCAD indicates that even though DeepCAD can recognize the primitives and the number of extrusions from the input texts, it fails to correctly parameterize those primitives and the extrusions. Compared to DeepCAD, the Text2CAD transformer predicts more accurate sketches and extrusion parameters. This can be observed in Figure ~\ref{fig:qual_eval}. 

\noindent Table~\ref{tab:quan_eval} (rows 2 and 3) shows the results between \twal and our final model. The incorporation of the Adaptive layer in the Text2CAD transformer improves the F1 scores for primitives such as lines, arcs, and circles. 
Notably, there is a $2.85\%$ increase of F1 scores for lines, $32.56\%$ for arcs, and $3.93\%$ for circles. Moreover, the improvement is particularly striking in terms of the IR, which sees a remarkable reduction by a factor of $\sim2.9$.

\noindent \textbf{B. GPT-4V Evaluation:}  To perform the \textit{visual inspections} of the 3D CAD models generated from abstract (L0), beginner (L1), intermediate (L2), expert (L3) level prompts, \gpt~\cite{achiam2023gpt} has been used. We follow the protocol outlined in~\cite{wu2023gpteval3d} and generate a meta prompt that consists of multi-view images of the reconstructed CAD models from both DeepCAD~\cite{Wu_2021_ICCV} and our model as well as the input text prompts. Following this, \gpt provides a verdict on which model predictions accurately reflect the text descriptions in terms of shape and geometry. But if the two models are very similar in shape and match the input text accurately, then it outputs \lq Undecided \rq.

\noindent We randomly select 1,000 samples for each level from the test dataset and generate a verdict per sample using \gpt~\cite{achiam2023gpt}. Table~\ref{tab:gpt4_human_eval} (\textit{left}) presents the final results. These results indicate that overall Text2CAD outperforms baseline DeepCAD~\cite{Wu_2021_ICCV}. Additionally, we observe that the performance gap between Text2CAD and DeepCAD is minimal at the abstract (L0) and beginner (L1) levels, despite Text2CAD losing by $2.8\%$ at the beginner level. However, as the complexity and parametric details of the prompts increase, the gap widens significantly. Text2CAD outperforms DeepCAD at the intermediate (L2) and expert (L3) levels, leading by as much as $18.6\%$ and $27.18\%$ respectively. This indicates that Text2CAD is more inclusive of the highly detailed design prompts. In supplementary Figure~\ref{fig:gpt_4_eval}, we have provided some examples.

%
%
%
%



\begin{table}[t]
\caption{We conduct GPT-4 evaluation of the CAD models generated from $1000$ prompts per level and User studies of $100$ samples per level. In both evaluations, overall Text2CAD is a favored choice over DeepCAD.}
\resizebox{\linewidth}{!}{
\begin{tabular}{ccccc|cccc}
\hline
\multirow{2}{*}{Model}        & \multicolumn{4}{c|}{GPT-4 Evaluation (\%)}                                                                                                                                                                                                                           & \multicolumn{4}{c}{User Study-based Evluation (\%)}                                                                                                                                                                                                                                \\
                              & \begin{tabular}[c]{@{}c@{}}Abstract Level\\ (L0)\end{tabular} & \begin{tabular}[c]{@{}c@{}}Beginner\\ Level (L1)\end{tabular} & \begin{tabular}[c]{@{}c@{}}Intermediate\\ Level (L2)\end{tabular} & \begin{tabular}[c]{@{}c@{}}Expert\\ Level (L3)\end{tabular} & \begin{tabular}[c]{@{}c@{}}Abstract Level\\ (L0)\end{tabular} & \begin{tabular}[c]{@{}c@{}}Beginner\\ Level (L1)\end{tabular} & \begin{tabular}[c]{@{}c@{}}Intermediate\\ Level (L2)\end{tabular} & \begin{tabular}[c]{@{}c@{}}Expert\\ Level (L3)\end{tabular} \\ \hline
\multicolumn{1}{l}{Undecided} & 0.80& 0.5                                                           & 1                                                                 & 0.70& -                                                             & -                                                             & -                                                                 & -                                                           \\
DeepCAD                       & 47.40& \textbf{51.15}& 40.20& 36.06& \textbf{50.95}                                                & 48.73                                                         & 44.94                                                             & 41.14                                                       \\
Text2CAD                      & \textbf{51.80}& 48.35& \textbf{58.80}& \textbf{63.24}& 49.05                                                         & \textbf{51.27}                                                & \textbf{55.06}                                                    & \textbf{58.86}                                    \\ \hline       
\end{tabular}
}
\vspace*{-.6\baselineskip}
\label{tab:gpt4_human_eval}
\vspace*{-.6\baselineskip}
\end{table}

\noindent \textbf{C. User Study-based Evaluation. } We conduct a user study with 100 randomly selected examples per level to evaluate the preference between our method and DeepCAD~\cite{Wu_2021_ICCV}. Five CAD designers with varying experience levels participate in the evaluation. Each participant is shown multi-view images of the reconstructed 3D CAD models from both methods side by side, along with their corresponding input texts. They are then asked to determine which CAD model is more geometrically accurate and easier to edit to achieve the desired outcome. Each participant evaluates $20$ samples per level. The final result is provided in Table~\ref{tab:gpt4_human_eval} (\textit{right}). To our surprise, the result follows a similar pattern as \gpt evaluation with a minimal performance gap in the abstract (L0) and beginner (L1) levels and a wider gap for more detailed intermediate (L2) and expert (L3) prompts.
\section{Limitation} \label{sec:limitation}
\vspace*{-0.3cm}
Despite the promising results of Text2CAD, several limitations exist. Firstly, LLaVA-NeXT~\cite{liu2024llava} is sensitive to the perspective distortions in multi-view images, which affects the generated shape descriptions and final LLM-generated prompts. Secondly, the lack of standardized benchmarks for evaluating text-to-CAD generation poses challenges in assessing model performance comprehensively. Furthermore, the DeepCAD~\cite{Wu_2021_ICCV} dataset is imbalanced, predominantly featuring rectangular and cylindrical shapes, which limits the model's robustness towards more complex shapes. Some failure cases are described in supplementary Section~\ref{sec:failure_cases}.
\section{Conclusion}
\vspace*{-.8\baselineskip} \label{sec:conclusion}
In this paper, we introduce Text2CAD, the first AI framework designed to generate parametric CAD models from text prompts suitable for users of all skill levels. Our contributions include a two-stage data annotation pipeline using Mistral-50B and LLaVA-NeXT and a novel end-to-end trainable Text2CAD Transformer architecture that effectively transforms natural language instructions into sequential CAD models. Through a comprehensive evaluation, including GPT-4V assessments and user studies by CAD designers, we demonstrate that Text2CAD outperforms the existing two-stage baseline, especially as the complexity and detail of the prompts increase. Future work will focus on addressing the current limitations, such as reducing annotation inaccuracies and improving dataset diversity, to further enhance the robustness and applicability of Text2CAD. 
\section{Acknowledgement} \label{sec:acknowledgement}
\vspace*{-.7\baselineskip} 
This work was in parts supported by the EU Horizon Europe Framework under grant agreement 101135724 (LUMINOUS).

\newpage
{
\bibliographystyle{plain} 
\bibliography{main}
}
\newpage

\section{CAD Sequence Representation} \label{sec:cad_seq_rep}
Table~\ref{tab:token} shows all the tokens used in our CAD sequence representation. We use the same representation as proposed by Khan et al~\cite{mkhan2024cadsignet} which uses a \textit{sketch-and-extrude} format. Each 2D sketch consists of multiple faces and each face consists of multiple loops and every loop either contains a line and a arc or a circle. Loops are always closed (\ie same start and end coordinate). We parameterize the curves in the following way
\begin{itemize}
    \item \textbf{Line:} Start and End coordinate
    \item \textbf{Arc:} Start, Mid and End coordiate
    \item \textbf{Circle:} Center and top-most coordinate
\end{itemize}
Finally, we represent a sketch using a sequence of 2D coordinates only with specific end tokens for the end of curve, loop, face and sketch. Each extrusion sequence consists of the $10$ parameters followed by an end of extrusion token. These are
\begin{itemize}
    \item \textbf{Euler Angles:}  $3$ parameters ($\theta, \phi, \gamma$) determining the orientation of the sketch plane.
    \item \textbf{Translation Vector:} $3$ parameters ($\tau_x,\tau_y, \tau_z$)  that describe the translation of the sketch plane.
    \item \textbf{Sketch Scale:} $1$ parameter ($\sigma$) for scaling the 2D sketches.
    \item \textbf{Extrude distances:}  $2$ parameters ($d^{+},d^{-}$) containing the extrusion distances towards and opposite of the normal of the sketch plane.
    \item \textbf{Boolean Operation:} $1$ parameter ($\beta$) determining the extrusion operation. There are $4$ extrusion operation in DeepCAD~\cite{Wu_2021_ICCV} dataset namely - \textit{solid body, cut, join and intersection}.
\end{itemize}
Except the boolean operation and all the end tokens, all the 2D sketch parameters as well as the extrusion parameters are quantized in $8$ bits.

\begin{table}[H]
    \caption{CAD sequence representation used in our experiment.}
    \resizebox{\linewidth}{!}{\begin{tabular}{ccccc}
    \hline
    \begin{tabular}[c]{@{}c@{}}Sequence\\ Type\end{tabular}                        & \begin{tabular}[c]{@{}c@{}}Token\\ Type\end{tabular} & \begin{tabular}[c]{@{}c@{}}Token\\ Value\end{tabular} & \begin{tabular}[c]{@{}c@{}}Token\\ Representation\end{tabular} & Description                                                                                \\ \hline
    \multicolumn{1}{l}{}                                                           & $pad$                                                & $0$                                                    & $(0,0)$                                               & Padding Token                                                                              \\
    \multicolumn{1}{l}{}                                                           & $cls$                                                & $1$                                                    & $(1,0)$                                               & Start Token                                                                                \\
    \multicolumn{1}{l}{}                                                           & $end$                                                & $1$                                                    & $(1,0)$                                               & End Token                                                                                  \\ \hline
    \multicolumn{1}{l}{\multirow{5}{*}{}}                                          & $e_s$                                                & $2$                                                    & $(2,0)$                                               & End Sketch                                                                                 \\
    \multicolumn{1}{l}{}                                                           & $e_f$                                                & $3$                                                    & $(3,0)$                                               & End Face                                                                                   \\
    \multicolumn{1}{l}{}                                                           & $e_l$                                                & $4$                                                    & $(4,0)$                                               & End Loop                                                                                   \\
    \multicolumn{1}{l}{}                                                           & $e_c$                                                & $5$                                                    & $(5,0)$                                               & End Curve                                                                                  \\
    \multicolumn{1}{l}{}                                                           & $(p_x,p_y)$                                          & $\Iintv{11,266}^2$                                    & $(p_x,p_y)$                                           & Coordinates                                                                                \\ \hline
    \multirow{11}{*}{\begin{tabular}[c]{@{}c@{}}Extrusion\\ Sequence\end{tabular}} & $d^+$                                                & $\Iintv{11,266}$                                      & $(d^+,0)$                                                          & \begin{tabular}[c]{@{}c@{}}Extrusion Distance Towards \\ Sketch Plane Normal\end{tabular}  \\ \cline{2-4}
                                                                                   & $d^-$                                                & $\Iintv{11,266}$                                       & $(d^-,0)$                                                         & \begin{tabular}[c]{@{}c@{}}Extrusion Distance Opposite \\ Sketch Plane Normal\end{tabular} \\ \cline{2-4}
                                                                                   & $\tau_x$                                             & $\Iintv{11,266}$                                       & $(\tau_x,0)$                                                      & \multirow{3}{*}{Sketch Plane Origin}                                                       \\
                                                                                   & $\tau_y$                                             & $\Iintv{11,266}$                                       & $(\tau_y,0)$                                                      &                                                                                            \\
                                                                                   & $\tau_z$                                             & $\Iintv{11,266}$                                       & $(\tau_z,0)$                                                      &                                                                                            \\ \cline{2-4}
                                                                                   & $\theta$                                             & $\Iintv{11,266}$                                       & $(\theta,0)$                                                      & \multirow{3}{*}{Sketch Plane Orientation}                                                  \\
                                                                                   & $\phi$                                               & $\Iintv{11,266}$                                       & $(\phi,0)$                                                        &                                                                                            \\
                                                                                   & $\gamma$                                             & $\Iintv{11,266}$                                       & $(\gamma,0)$                                                      &                                                                                            \\ \cline{2-4}
                                                                                   & $\sigma$                                             & $\Iintv{11,266}$                                       & $(\sigma,0)$                                                      & Sketch Scaling Factor                                                                      \\
                                                                                   & $\beta$                                              & $\{7,8,9,10\}$                                         & $(\beta,0)$                                                        & Boolean (New, Cut, Join, Intersect)                                                        \\
                                                                                   & $e_e$                                                & $6$                                                    & $(6,0)$                                                   & End Extrude                                                                                \\ \hline
    \end{tabular}}
    \label{tab:token}
    \vspace{-.8\baselineskip}
    \end{table}

\section{Implementation Details for Data Annotation}\label{sec:additional_details_data_annot}
As mentioned in Section~\ref{sec:text2cad_data_annot}, we use LLaVA-NeXT~\cite{liu2024llava} for the VLM oriented task and Mistral-50B~\cite{jiang2023mistral} for LLM tasks. We use $1$ Nvidia A$100$-$40$GB GPU to run LLaVA-NeXT and $4$ Nvidia A$100$-$80$GB GPUs to run Mistral-50B.

\section{Additional Details on Data Annotation Pipeline}\label{sec:additional_details_data_annot_2}
\begin{figure}[t]
    \centering
    \includegraphics[width=1.0\linewidth]{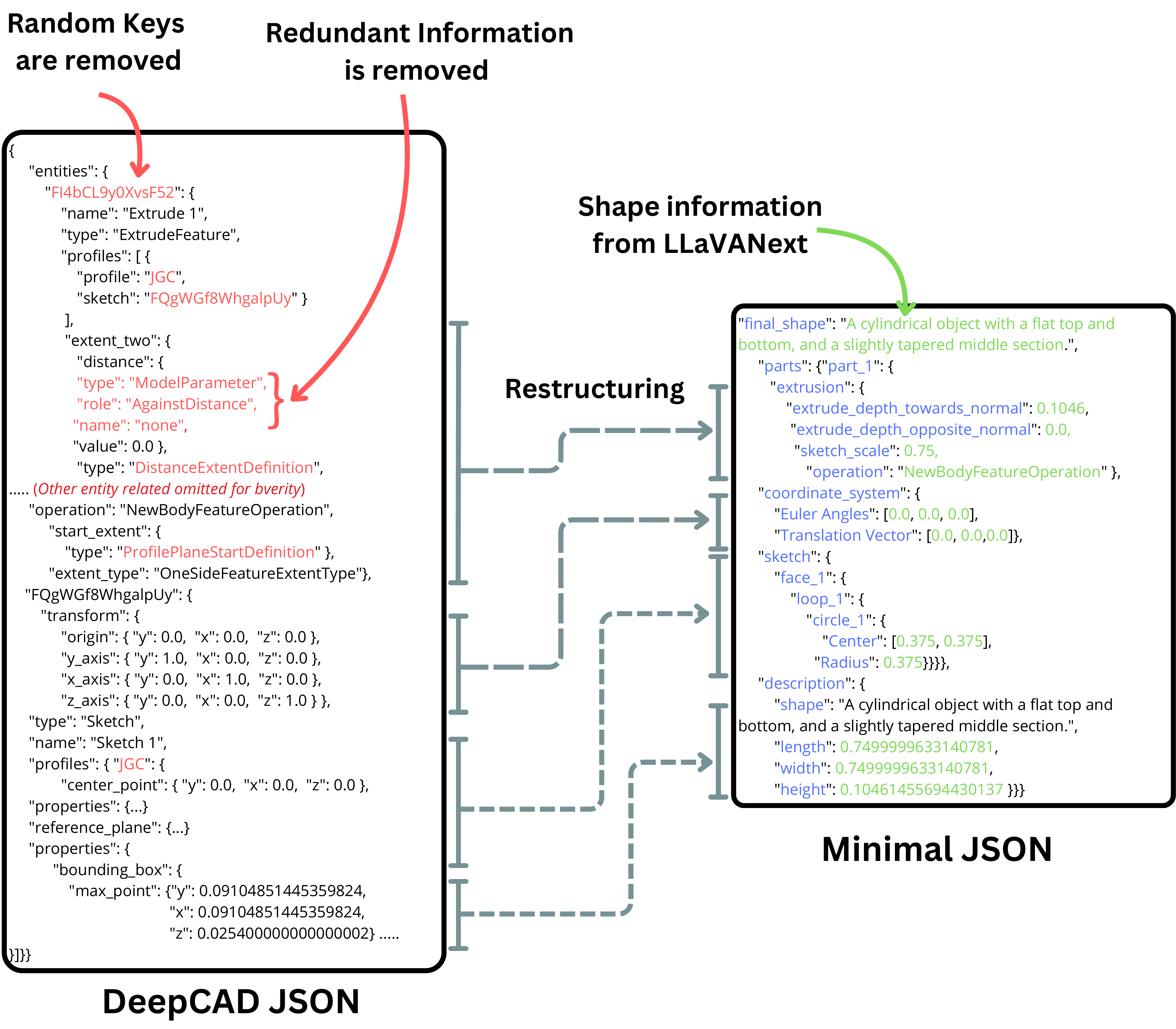}
    \caption{An example of Minimal metadata JSON (\textit{right}) generated from DeepCAD~\cite{Wu_2021_ICCV} JSON (\textit{left}). During the minimal metadata generation, random keys (\eg \textit{"FI4bCL9y0XvsF52"}) or redundant design information (\eg \textit{\{"type": "ModelParameter", "role": "AgainstDistance"\}}) is removed.}
    \label{fig:raw2json}
\end{figure}

Since LLMs are prone to hallucinations~\cite{llm_hallucinations}, we employ several strategies to mitigate this issue. Firstly, we observe that when we directly pass the raw DeepCAD jsons to Mistral instead of the minimal metadata to generate the detailed natural language instructions, the model often uses the random keys provided in the dictionary to refer to the corresponding curves or the sketches. Additionally, the json contains redundant information that is not necessary in the final CAD construction. To overcome this, we generate minimal metadata by restructuring the dictionary into a more human-readable format. In Figure~\ref{fig:raw2json}, we provide an example DeepCAD Json (\textit{left}) and a minimal data(\textit{right})

\section{Discussion on Prompt Diversity}\label{sec:prompt_diversity}
\begin{figure}[H]
    \centering
    \includegraphics[width=0.7\linewidth]{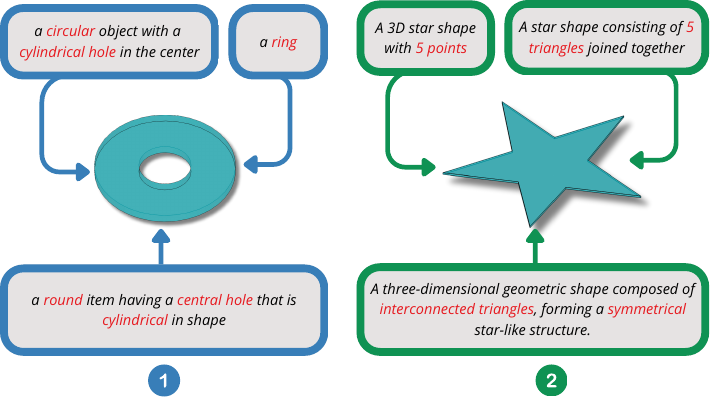}
    \caption{Visual examples of 3D CAD model generation using varied prompts. \textcolor{blue}{\textbf{(1)}} Three different prompts yielding the \textit{same ring-like model}, some \textit{without explicitly mentioning 'ring'}. \textcolor{darkgreen}{\textbf{(2)}} Three diverse prompts resulting in \textit{same star-shaped model}, each emphasizing \textit{different star characteristics}.}
    \label{fig:diversity}
\end{figure}

The diversity of the generated textual prompts in our Text2CAD dataset depends on variety of CAD models available in the existing dataset and the performance of Mistral and LLaVA-Next. As mentioned in Section~\ref{sec:limitation}, DeepCAD dataset which is the only large-scale dataset containing full design history, is imbalanced towards simpler shapes, featuring a lot of rectangular and cylindrical shapes. Thus to increase the diversity in textual prompts, we have focused more on generating shape descriptions from LLaVA-Next rather than only object names. For example, in our annotation \textit{"a ring"} can be sometimes described as \textit{"a circular object with a cylindrical hole in the center"}. This approach enables our transformer model to learn to generate the same CAD models using different styles of textual prompts. In Figure~\ref{fig:diversity}, we show two examples where the network generated same CAD models from various types of prompts. 

\section{Additional Experimental Details on Interpolated promps}\label{sec:interpolation}
\begin{figure}
    \centering
     \includegraphics[width=1.0\linewidth]{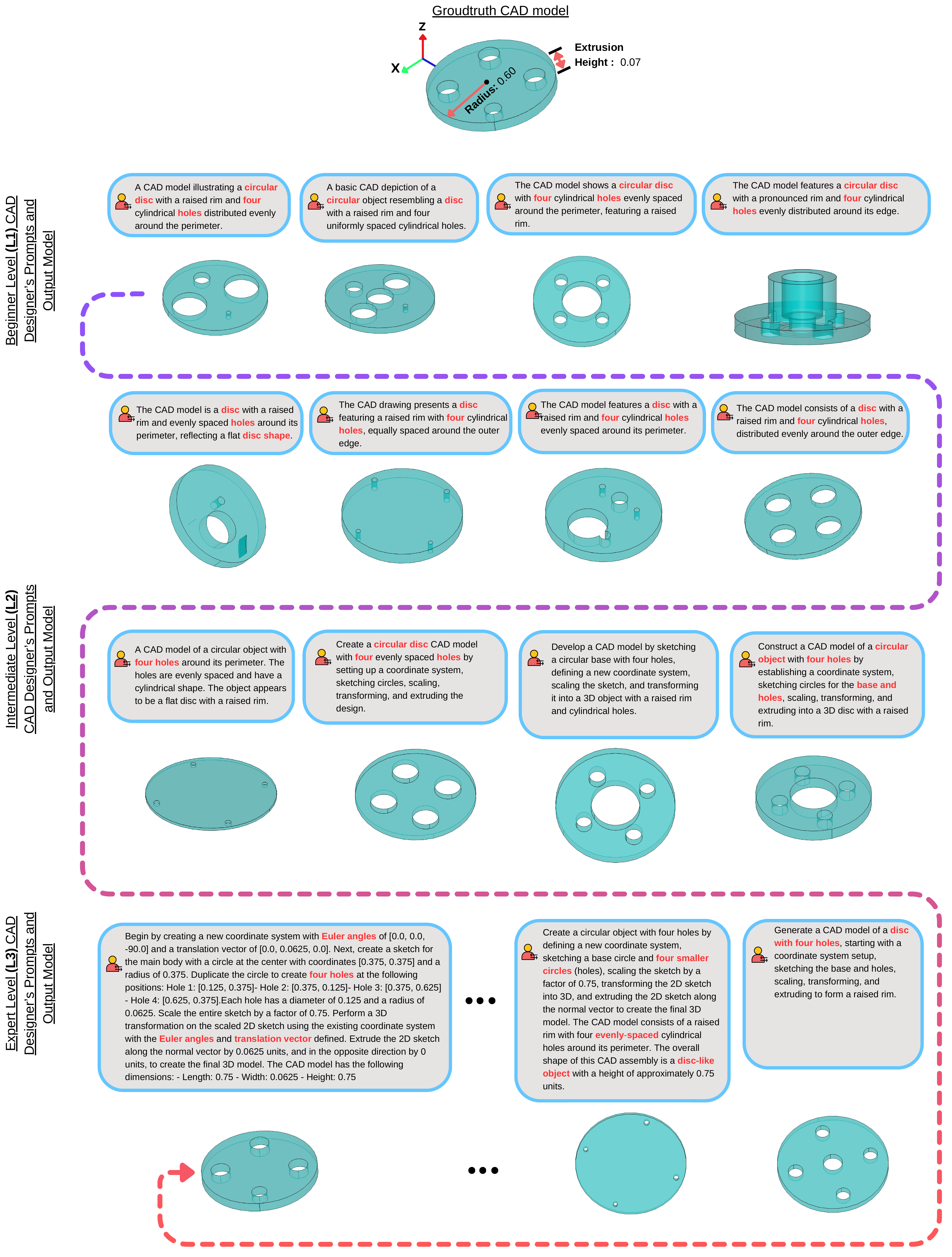}
    \caption{Visual results of Text2CAD on interpolated text prompts generated by GPT-4V. From top to bottom, the geometric details in the text prompts increase.}
    \label{fig:Beginner_to_Expert_1}
\end{figure}

\begin{figure}[]
    \centering
     \includegraphics[width=1.0\linewidth]{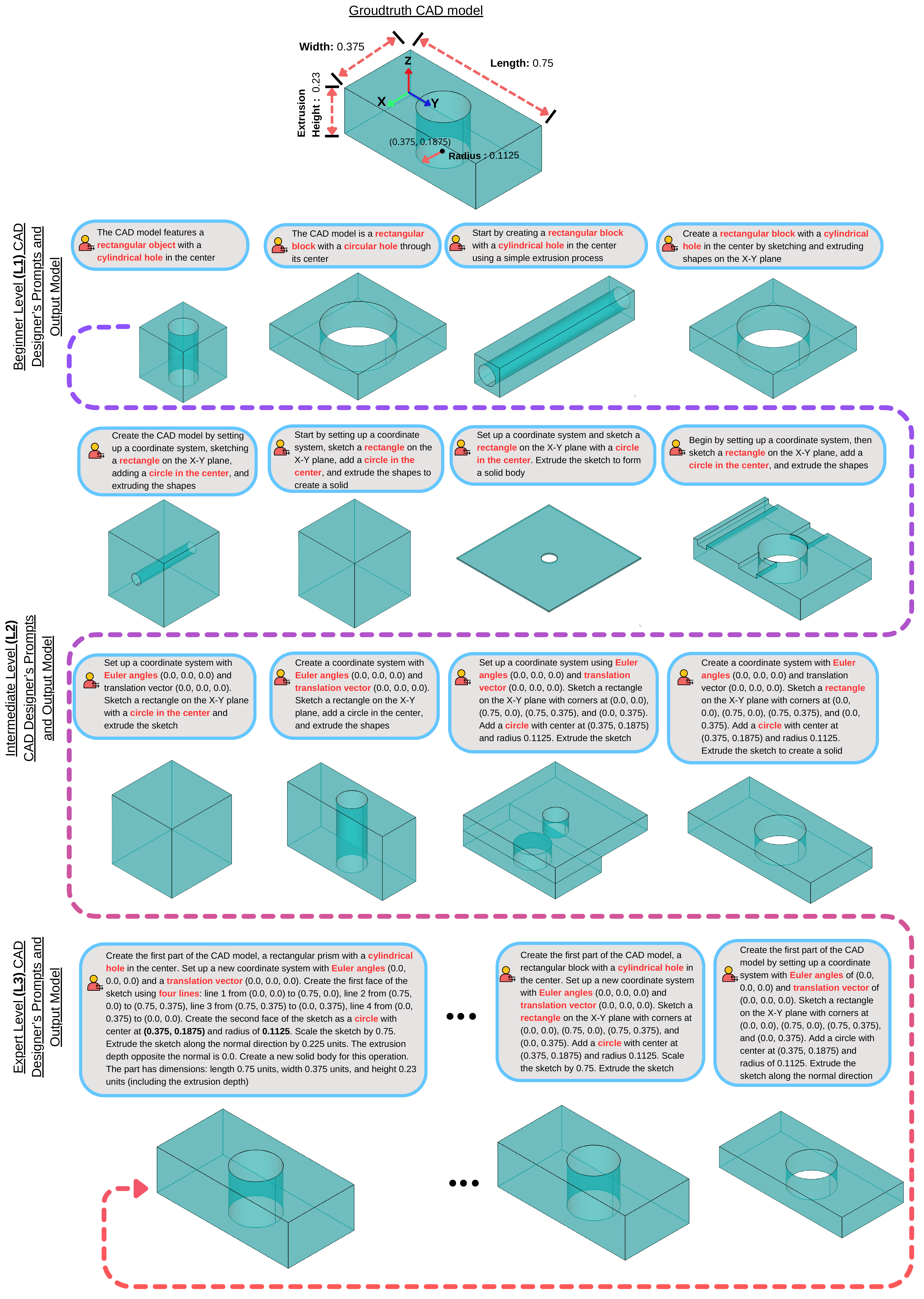}
    \caption{Visual results of Text2CAD on interpolated text prompts generated by GPT-4V. From top to bottom, the geometric details in the text prompts increase.}
    \label{fig:Beginner_to_Expert_2}
\end{figure}

In this section, we provide details on our model's performance on text prompts that contain different structure than the training samples. To generate these prompts, we pass all four text prompts (\ie abstract, beginner, intermediate and expert) of a CAD model to \gpt~\cite{achiam2023gpt} and ask to generate $20$ more samples by interpolating between the all the levels. Figure~\ref{fig:Beginner_to_Expert_1} and Figure~\ref{fig:Beginner_to_Expert_2} shows visual results of two examples. The results indicate that Text2CAD can effectively handle varying level of prompt structures, generating accurate and similar CAD models in shape compared to the ground truth. It retains the overall geometry for prompts with highly abstract shape descriptions (\textit{first} and \textit{second} row in Figure~\ref{fig:Beginner_to_Expert_1} and Figure~\ref{fig:Beginner_to_Expert_2} ) . As the parametric details increase in the text prompts, it generates the precise CAD model as the ground truth (\textit{third} and \textit{fourth} row in Figure~\ref{fig:Beginner_to_Expert_1} and Figure~\ref{fig:Beginner_to_Expert_2})

\section{Additional Qualitative Samples}\label{sec:add_qual_samples}
\begin{figure}[t]
    \centering
    \includegraphics[width=1.0\linewidth]{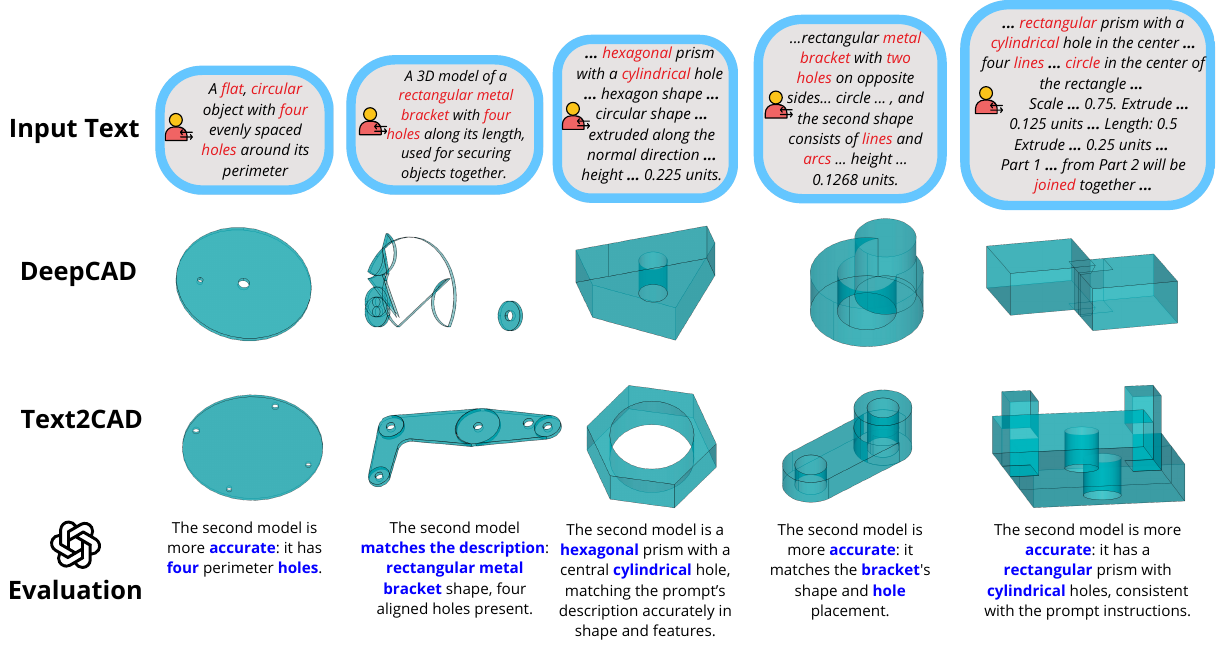}
    \caption{Additional qualitative results of the reconstructed CAD models of DeepCAD~\cite{Wu_2021_ICCV} and Text2CAD on DeepCAD~\cite{Wu_2021_ICCV} dataset. From top to bottom - Input Texts, Reconstructed CAD models using DeepCAD and Text2CAD respectively and \gpt Evaluation.}
    \label{fig:additional_qual}
\end{figure}

Some additional qualitative samples are presented in Figure~\ref{fig:additional_qual}.

\section{Discussion on Failure Cases}\label{sec:failure_cases}

\begin{figure}
    \centering
    \includegraphics[width=1.0\linewidth]{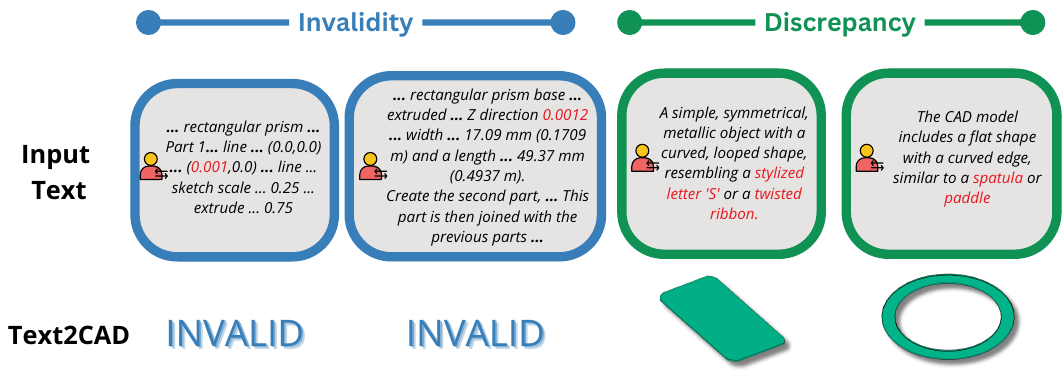}
    \caption{Failure cases for Text2CAD Transformer. \textbf{Invalid Samples} (left): The network fails to generate any valid CAD model. \textbf{Discrepancy Cases} (right): The generated CAD model does not match the input text prompts.}
    \label{fig:failure_cases}
\end{figure}

In this section, we describe two types of failure cases for Text2CAD Transformer. In Figure~\ref{fig:failure_cases}, we have shown examples of both type of cases.

\textbf{1. Invalidity}: In this scenario, the model fails to generate any CAD model from the text prompts. As reported in Table~\ref{tab:full_eval}, this occurs in approximately $1\%$ of the test samples. In these cases, the model predicts invalid sketch or extrusion parameters, such as the same start and end points for lines or arcs, or zero values for extrusion depth on both sides.

\textbf{2. Discrepancy}: Discrepancy refers to situations where the generated model does not precisely match the shape described in the text prompts. This is more prevalent in our model than invalidity and is harder to quantify. We notice that this occurs when prompts are more focused on object name (\eg spatula, paddle) rather than parametric descriptions. We argue that this issue comes from noisy VLM annotations. As mentioned in the Section~\ref{sec:limitation}, the perspective distortions in multi-view images can affect the accuracy of shape or object name recognition. These noisy descriptions propagate into other prompt levels using the annotation pipeline.

%
\begin{figure}[t]
    \centering
    \includegraphics[width=1.0\linewidth]{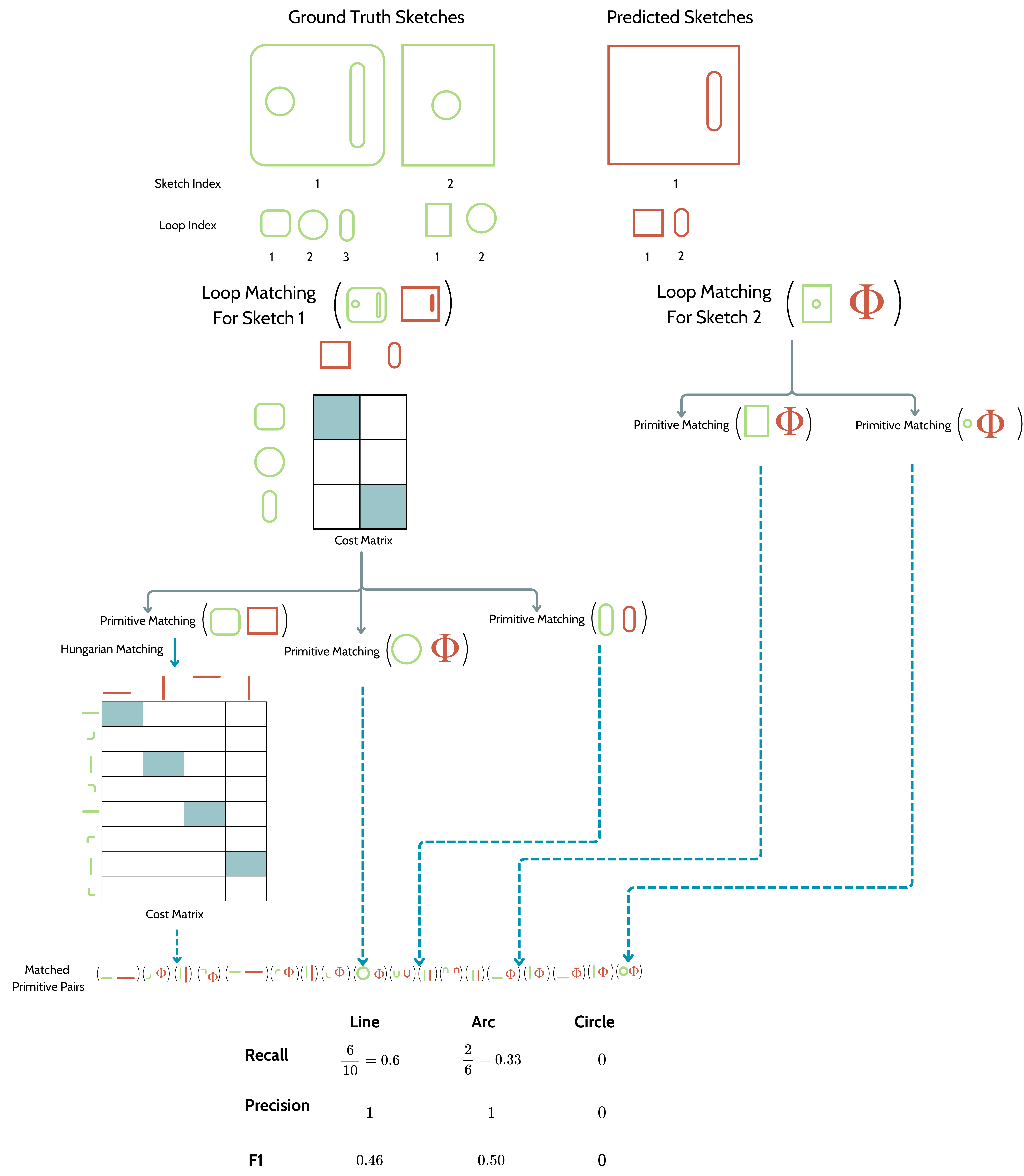}
    \caption{F1 score calculation for CAD sequence evaluation as proposed in~\cite{mkhan2024cadsignet}.}
    \label{fig:loop_matching}
\end{figure}

\newpage

\begin{figure}[t]
    \centering
    \includegraphics[width=1.0\linewidth]{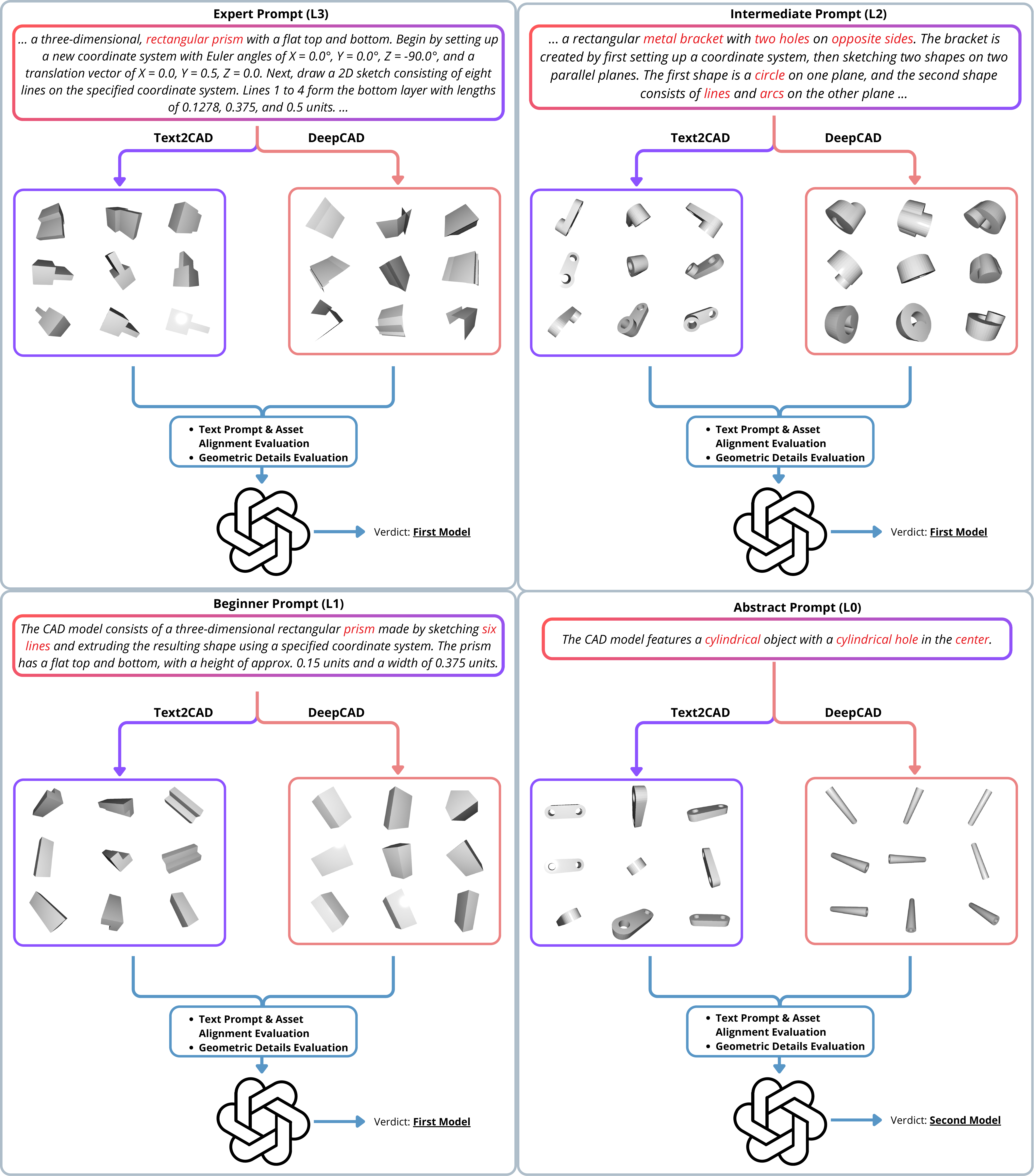}
    \caption{\textbf{GPT-4V Evaluation Strategy}: Four prompts (one per level) are randomly sampled from the test set. These prompts are used to reconstruct parametric CAD models from the predicted CAD sequences using both DeepCAD and the proposed Text2CAD. Multi-view images of these models are generated, which are used for the \gpt evaluation. \gpt analyzes their alignment with the initial text prompt and their geometric details and provides a final verdict for this comparison. As shown in the image, our model performs better when input text prompts contain more parametric details.}
    \label{fig:gpt_4_eval}
\end{figure}

\vspace*{.6\baselineskip}

\pagebreak

\end{document}